\documentclass[sigconf]{acmart}
\usepackage{subcaption}
\acmSubmissionID{14330.7}

\AtBeginDocument{%
  \providecommand\BibTeX{{%
    \normalfont B\kern-0.5em{\scshape i\kern-0.25em b}\kern-0.8em\TeX}}}

\copyrightyear{2024}
\acmYear{2024}
\setcopyright{acmlicensed}
\acmConference[SAP '24]{ACM Symposium on Applied Perception 2024}{August 30--31, 2024}{Dublin, Ireland}
\acmBooktitle{ACM Symposium on Applied Perception 2024 (SAP '24), August 30--31, 2024, Dublin, Ireland}
\acmDOI{10.1145/3675231.3675238}
\acmISBN{979-8-4007-1061-2/24/08}
\begin{document}

\title[Estimating roadway billboard salience]{A method for estimating roadway billboard salience}

 \author{Zuzana Berger Haladova}
 \orcid{0000-0002-5947-8063}
\email{zuzana.bergerhaladova@fmph.uniba.sk}
 \affiliation{
   \institution{Faculty of Mathematics Physics and informatics, Comenius University}
   \city{Bratislava}
  \country{Slovakia}
 }

 \author{Michal Zrubec}
\orcid{0009-0005-8024-6645}
 \email{m.zrubec60@gmail.com}
 \affiliation{
   \institution{Faculty of Mathematics Physics and informatics, Comenius University}
\city{Bratislava}
    \country{Slovakia}
\email{m.zrubec60@gmail.com}
}

 \author{Zuzana Cernekova}
 \orcid{0000-0002-7617-4192}
 \email{cernekova@fmph.uniba.sk}
 \affiliation{%
  \institution{Faculty of Mathematics Physics and informatics, Comenius University}
\city{Bratislava}
  \country{Slovakia}
  }


\begin{abstract}
Roadside billboards and other forms of outdoor advertising play a crucial role in marketing initiatives; however, they can also distract drivers, potentially contributing to accidents. This study delves into the significance of roadside advertising in images captured from a driver's perspective. Firstly, it evaluates the effectiveness of neural networks in detecting advertising along roads, focusing on the YOLOv5 and Faster R-CNN models. Secondly, the study addresses the determination of billboard significance using methods for saliency extraction. The UniSal and SpectralResidual methods were employed to create saliency maps for each image. The study establishes a database of eye tracking sessions captured during city highway driving to assess the saliency models.
\end{abstract}





\maketitle

\section{Introduction}



Outdoor advertising, particularly through roadside billboards, plays an integral role in contemporary marketing strategies, serving as a prominent medium for brand promotion. However, the widespread presence of these advertisements along roadways raises concerns about potential distractions for drivers, which could lead to safety hazards and accidents. Recognizing the dual impact of outdoor advertising on marketing effectiveness and road safety, this study investigates the nuanced significance of roadside advertising through a driver's perspective.

This study leverages the capabilities of neural networks. The YOLOv5 and Faster R-CNN models are evaluated for their effectiveness in detecting advertising spaces along roads. Furthermore, the study delves into the determination of billboard significance through saliency extraction methods. By employing the UniSal and SpectralResidual techniques, the research aims to create saliency maps for each image, providing insights into the key features that capture a driver's attention. 

To ground the findings in real-world driving scenarios, the database of eye tracking sessions conducted during city highway driving was created within the study. By utilizing eye tracking data, the research seeks to evaluate the effectiveness of the saliency models in capturing the attention of drivers.


The paper is organized as follows. The second section provides an overview of the previous works in the area of billboard/object detection and saliency extraction as well as the research on the relevance and viewability of billboards. In the third section datasets used for training and evaluation of the proposed methods are discussed. In the fourth section, the proposed method is described. In the fifth section, the results of the acquired experiments are presented. Then the Conclusions are given.

\section{Related work}
This section delves into previous research on the relevance and viewability of billboards, billboard detection, and saliency extraction methods. In the area of billboard relevance, several factors such as gaze duration, placement, and content impact on driver attention and road safety were examined. The placement of advertising significantly affects its reach. Advertisements on the driver's side of the road attract more fixations than those on the opposite side \cite{Costa19}. Contrary to the assumption that the height of the advertisement placement prolongs fixation duration, studies reveal that drivers look more at advertisements at road level than those elevated three or more meters above the ground, except when drivers intentionally seek out higher-placed advertisements. Additionally, factors such as the angle of the advertising surface, the complexity of the traffic section, and the environment influence the visibility of advertisements \cite{Zalesinska18,Mollu18,Costa19,Crundall06}.

Content is another crucial characteristic. Research shows that advertisements with longer text, a sexual undertone, or featuring human beings result in longer fixations \cite{Harasimczuk21, Meuleners20,Maliszewski19,Tarnowski17}. Moreover, more complex advertisements tend to capture the driver's gaze for a longer duration \cite{Marciano17}. Advertisements eliciting negative emotions in drivers also lead to longer fixations and reduced speed, with negative content being more memorable according to post-drive questionnaires filled out by drivers \cite{Chan13}. 
While our study focuses solely on the saliency of billboards, it can be extended in the future to analyze other important aspects that increase the duration of fixations among drivers.

\subsection{Billboard detection}

Since billboard detection is a special case of object detection, we will discuss it further in the following section. Convolutional Neural Networks (CNNs) for object detection typically use a universal backbone for extracting image features, paired with a framework to recognize object classes and generate bounding boxes. Two primary categories of object detectors have emerged: single-stage and two-stage detectors.

In two-stage detectors, a network first proposes objects, which are then verified in a second stage. The initial bounding boxes are refined in this second stage for greater precision. Faster R-CNN is a notable example of a two-stage detector, known for its accuracy \cite{FasterRCNN}. Its extension, Mask R-CNN, also detects object masks \cite{MaskRCNN}.

Single-stage detectors refine bounding boxes in one pass through the network. This is exemplified by anchor-based methods like RetinaNet \cite{RetinaNet} and anchor-free approaches like YOLO \cite{Yolo} and FCOS \cite{tian2019fcos}. Keypoint-based methods detect key points at bounding box corners \cite{CornerNet} or centers \cite{CenterNet}, with regression outputs to define the boxes. ATSS offers a unique single-stage approach with adaptive training sample selection \cite{zhang2020bridging}.


There were also several studies dealing with the CNN methods for outdoor advertising detection, usually relying on transfer learning \cite{chavanbillboard, hossari2018adnet, rahmat2019advertisement, Liu2018}. For instance, one approach \cite{rahmat2019advertisement} used AlexNet’s Deep Convolutional Neural Network (DCNN) with inductive transfer learning, achieving high training accuracy for advertisement billboard detection. Other methods, such as the attention-based multi-scale feature fusion region proposal network (AM-RPN) based on Faster R-CNN \cite{Liu2018}, incorporated advanced techniques like adaptive non-maximum suppression to enhance detection performance in challenging conditions. 

\subsection{Saliency extraction}
Much of the existing literature on visual saliency modeling focuses on predicting human visual attention mechanisms in static scenes. First models \cite{itti1998model, sun2003object, judd2009learning, hou2007saliency} primarily concentrate on low-level image features, such as intensity, contrast, color, and edges, referred to as bottom-up methods. One of them the Spectral Residue \cite{hou2007saliency} is considered fast and robust.
Recent advancements leverage deep neural networks, starting with Vig et al. \cite{vig2014large}. Jiang et al. \cite{jiang2015salicon} collected SALICON, a large-scale saliency dataset, facilitating exploration in deep learning-based saliency modeling. Other studies center around network architecture design, and varying model sizes. In \cite{droste2020unified} Droste et al. proposed UNISAL saliency extraction method achieves state-of-the-art performance on all video saliency datasets and was on par with the state-of-the-art for image saliency datasets, despite faster runtime and a 5 to 20-fold smaller model size compared to all competing deep methods. For a thorough review of saliency extraction methods see \cite{abraham2023visual, ullah2020brief}. 

\section{Dataset}
This section discuss datasets used and created within this work. 

In our solution, we use two datasets. The first dataset is the existing Mapillary Vistas Dataset \cite{neuhold2017mapillary}, which is ideal due to its extensive street scenes and diverse object annotations. It contains 25,000 images with annotations for training and validation sets, divided into 18,000 training images, 2,000 validation images, and 5,000 test images.

\begin{figure}[]
   \captionsetup[subfigure]{justification=centering}
   \centering  
     \includegraphics[width=0.4\textwidth]{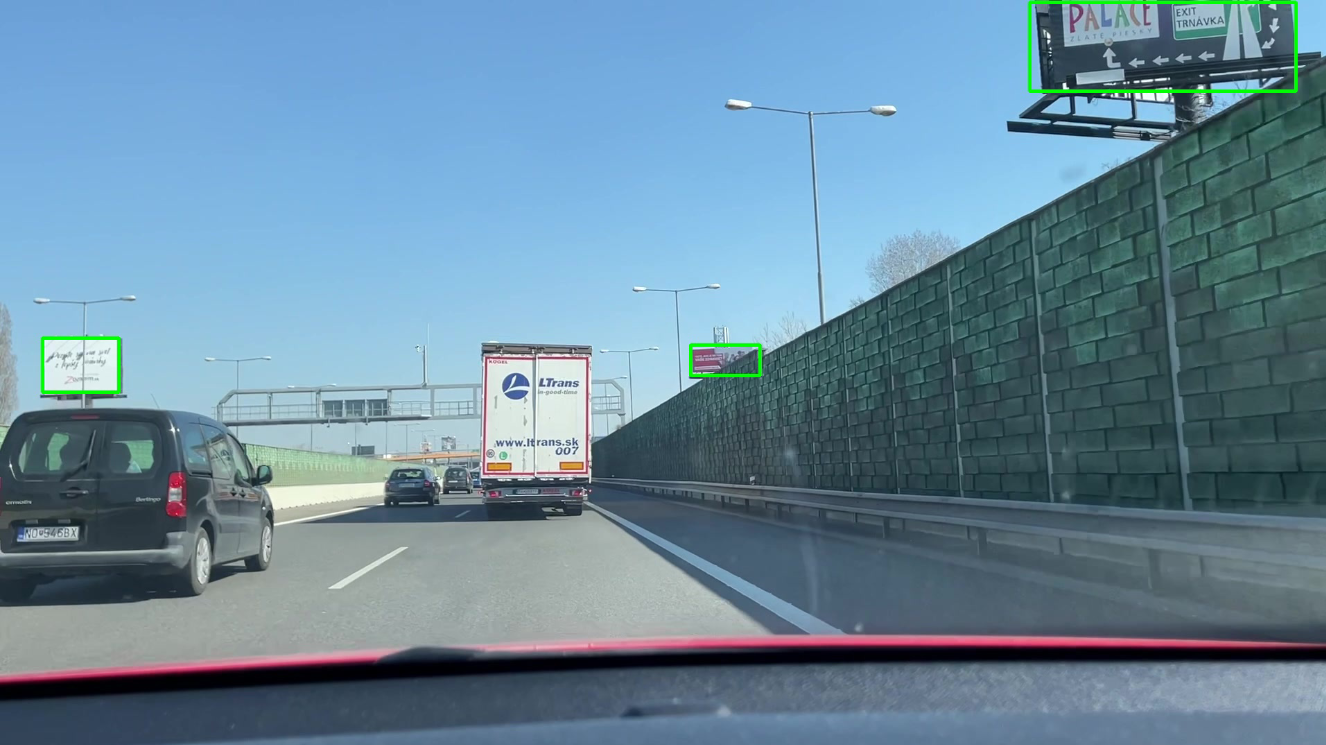}
     \caption{Image with annotated billboards from our dataset}
     \label{fig:annotation}
   \end{figure}

The second dataset is custom, created using car dashboard imagery and an eyetracker. For eyetracking, we used the Tobii Pro Glasses 3. The roads were chosen to represent various driving scenarios, including highways and urban streets, aiming to cover a wide range of real-world conditions. This dataset has 1580 images, standardized to $1920 \times 1080$ pixels. Figure \ref{fig:annotation} shows an example.

Creating this dataset was challenging and time-consuming, requiring many images. All images were manually annotated, with each image having one associated file. We used Roboflow to label the billboards, focusing on large advertisements and excluding smaller ads, such as store names and distant billboards. This dataset will be available online (link omitted due to double-blind review). It was created with images from 3 drivers (2 males, 1 female, ages 22-26) and divided into training (1350 images), validation (150 images), and test (80 images) sets for model training and evaluation.


\section{Proposed method}
\begin{figure*}
    \centering
    \includegraphics[width=0.8\textwidth]{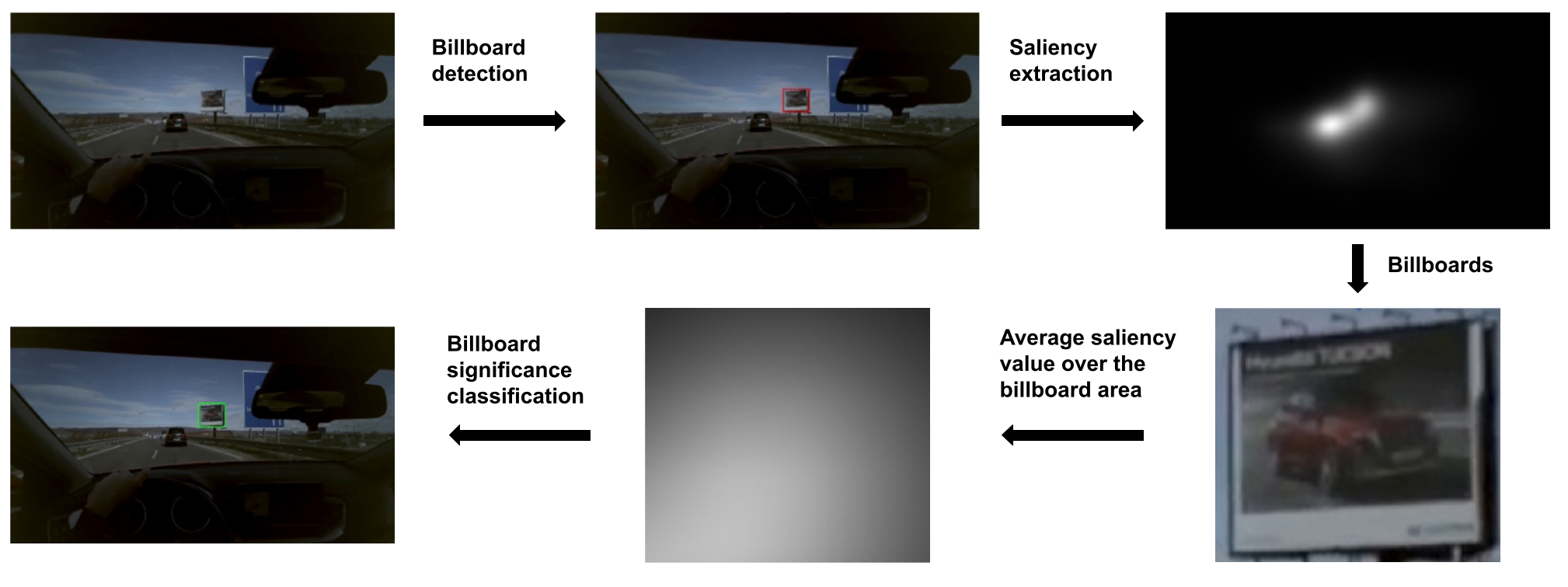}
    \caption{The scheme of the process of classification of the significance of billboards on the images}
    \label{fig:escheme}
\end{figure*}

To address the challenge of determining the saliency of advertisements, we propose the following procedure.
\begin{enumerate}
    \item detection of the advertising areas in the input image,
    \item  creating a saliency map for the input image,
    \item obtaining the average saliency of the advertisement areas,
    \item  checking whether the average significance value of the ad space is large enough,
     \item classifying the saliency of the advertising spaces.
\end{enumerate}
A scheme of the process can be seen in Figure \ref{fig:escheme} and detailed explanation of steps is provided in the subsequent text.

To detect billboard areas, we opted to test the YOLO\cite{Yolo} and Faster R-CNN models\cite{FasterRCNN}. These models 
achieve very good results in terms of speed while maintaining sufficient detection accuracy. During the evaluation, we use models pretrained on the MS COCO dataset \cite{MSCOCO}. 
We evaluate the models and use the better model to detect the billboard areas for the following saliency detection.

\subsection{Saliency extraction methods}
After identifying regions of interest using object detection models, the next step involves generating a saliency map from the original image.  To achieve this, we explored two distinct methods designed for saliency extraction: spectral residue\cite{hou2007saliency} and UniSal \cite{droste2020unified}. The \textbf{Spectral Residue} method operates by breaking down the image into its phase and amplitude components using Fourier transform. By smoothing the amplitude component, we emphasize outliers, which are areas significantly different from the average. Subtracting this smoothed amplitude from the original component reveals regions that stand out the most, highlighting salient areas effectively \cite{hou2007saliency}. In contrast, the \textbf{UniSal} model \cite{droste2020unified} offers a more comprehensive approach. It is specifically engineered to handle saliency modeling in both images and videos. The architecture utilizes an encoder - RNN - decoder design tailored for salience modeling. The model first encodes the input image using MobileNet-V2 \cite{sandler2018mobilenetv2}, capturing its essential features. Then, a convolutional Gated Recurrent Unit (GRU) RNN processes this information to capture temporal dependencies, if applicable. This sophisticated design allows UniSal to effectively identify salient regions in diverse visual content

\subsection{Billboard Significance Assessment}
After obtaining the saliency maps, we determine the significance of billboards by assessing the average saliency value within the bounding box of each billboard area. Using the Spectral residual and UniSal methods, we generate saliency maps, focusing specifically on regions corresponding to advertising areas.
In these areas, we normalized the saliency values of all pixels to an interval from 0 to 1. The average saliency value in each area is obtained by summing and dividing these normalized values by the number of pixels in the area.
Similarly, we compute the average saliency value for all advertising areas using the same methodology. To classify the significance of an advertisement, we compared the average salience value of the advertising space and the average salience value of all advertising spaces. If the average salience value is greater, the advertising space is classified as salient. 
Subsequently, we validate the significance of billboards based on eyetracker data. A billboard is classified as significant if corresponding eye fixations are observed within the billboard area in the eyetracker image. This serves as a crucial validation step in confirming the impact and visibility of billboards.


\section{Results}
The evaluation was performed in two phases. At first, we focused on the evaluation and comparison of the YOLOv5 and Faster R-CNN models, which are used to detect advertising areas. Then, we evaluated the methods for extracting saliency in the image and the method for classifying the significance of advertising spaces.

For testing purposes, we used the test dataset that consists of 80 images containing a total of 210 billboard areas. The dataset contains images with a resolution of $1920 \times 1080$, featuring billboard areas from both highways and cities. These images include diverse lighting conditions, weather scenarios, and instances where billboard areas may be occluded with other objects.

For the evaluation of the model, we have employed standard metrics used to determine the accuracy of a detector, the Intersection over Union (IoU). It expresses how much the bounding box of the predicted object overlaps with the actual bounding box of the object.
Another widely used metric for evaluating object detectors is average precision (AP). It is used e.g. in COCO challenges. The average precision is evaluated over several IoU values ranging from $0.5$ to $0.95$ with a step of $0.05$. 

\begin{table}[ht]
\centering
\caption{Results of models for detecting billboards after fine-tuning on Mapillary Vistas dataset.}
\label{table:cat1}
\begin{tabular}{|l c c|}
 \hline
 Model &	AP@0.5 &	AP@0.5:0.95 \\ [0.5ex]
 \hline
YOLOv5 &	63.2\% &	48.1\% \\ [0.1ex]
Faster R-CNN &	55.8\% &	40.2\% \\ [0.1ex]
 \hline
\end{tabular} \label{obj_det_mapilary}
\end{table}
In Table \ref{obj_det_mapilary} we can see the results of the models after fine-tuning on Mapillary Vistas dataset. 
We experimented with various combinations of data augmentation techniques, and the most favorable outcomes were achieved when incorporating random rotation (ranging from $-10^{\circ}$ to $10^{\circ}$), random brightness adjustment (ranging from $-10\%$ to $10\%$), random Gaussian noise insertion, and random clipping.

Next, testing of the models fine-tuned on our dataset was performed. The results are shown in the Table \ref{obj_det_our}.
\begin{table}[ht]
\centering
\caption{Results of models for detecting billboards after fine-tuning on our dataset.}
\label{table:cat2}
\begin{tabular}{|l c c|}
 \hline
 Model &	AP@0.5 &	AP@0.5:0.95 \\ [0.5ex]
 \hline
YOLOv5 &	96.6\% &	69.5\% \\ [0.1ex]
Faster R-CNN &	89.6\% &	58.4\% \\ [0.1ex]
 \hline
\end{tabular} \label{obj_det_our}
\end{table}

In Tables \ref{obj_det_mapilary} and \ref{obj_det_our}, we can notice that significantly better results were obtained on our dataset than on the Mapillary Vistas dataset. The reason for these differences is that in the Mapillary Vistas dataset, the images are of a much higher resolution (at most up to $5248 \times 3936$ pixels), where even very small and distant advertising areas are annotated. When training the models for object detection, the size of the images is reduced to a smaller size ($1920\times 1080$ pixels in our case) on input, and thus the very small advertising areas annotated on the high-resolution images are "faded out".

\subsection{Comparing Saliency Maps to Fixation Maps}
We analyzed the saliency maps generated by our models against the ground truth fixation maps gathered using the eyetracker. To evaluate the consistency between the saliency and fixation maps, we utilized two evaluation metrics: Area Under ROC Curve (AUC) and Normalized Scanpath Saliency (NSS) \cite{Riche-13}. These metrics quantify the agreement between the predicted saliency and actual fixation points, providing complementary insights into the effectiveness of our saliency extraction methods.
 
The ideal value of the AUC-Judd metric is 1, which means that it represents a 100\% detection accuracy. The random saliency map has a resulting value of 0.5, which means that using our methods, we should obtain a value greater than 0.5. 

A positive NSS score indicates that the saliency value at a fixation point is above the mean saliency value, while a negative score indicates a saliency value below the mean. The mean used here is the average of all standardized saliency values at fixation points. For instance, an NSS score of 1 means that the saliency values at fixation points were 1 standard deviation above the mean saliency value. Unlike AUC, NSS works with actual saliency values and is more sensitive to false positives, providing a nuanced assessment of how well a saliency model aligns with human eye fixations.

\begin{table}[ht]
\centering
\caption{Results of saliency models.}
\label{table:cat3}
\begin{tabular}{|l c c|}
 \hline
 Method &	AUC &	NSS \\ [0.5ex]
 \hline
Unisal &	0,926 &	2,287 \\ [0.1ex]
SpectralResidual &	0,915 &	2,921 \\ [0.1ex]
 \hline
\end{tabular} \label{saliency_det_our}
\end{table}
In Table \ref{saliency_det_our} we can see that both used methods performed well and achieved good results on the test dataset.

\subsection{Significance of the billboards}

In assessing the significance of billboards, we computed the average saliency values for all advertising spaces in the training set. The resulting average significance value, normalized to the interval from 0 to 1, was determined to be 0.416.

\begin{table}[ht]
\centering
\caption{The billboards' significance results.}
\label{table:cat4}
\begin{tabular}{|l c c|}
 \hline
 Method &	Accuracy &	Sensitivity \\ [0.5ex]
 \hline
Unisal &	82,6\% &	74,0\% \\ [0.1ex]
SpectralResidual &	69,2\% &	78,3\% \\ [0.1ex]
 \hline
\end{tabular} \label{billboard_significance}
\end{table}
In Table \ref{billboard_significance}, we can observe that our method for detecting the significance of billboard areas achieved an accuracy of 82.6\% and a sensitivity of 74.0\% when generating saliency maps using the UniSal method. When employing the SpectralResidual method, the accuracy was 69.2\%, and the sensitivity was 78.3\%. This implies that when using the UniSal method, there is a higher likelihood that predicted positive detections were indeed correct.

\section{Conclusion and discusion}
In this study, we introduced a method for classifying the significance of billboards. 
Our evaluation indicates that YOLOv5 outperforms Faster R-CNN in the task of detecting billboard areas, demonstrating higher accuracy and speed. The best results were achieved after fine-tuning both YOLOv5 and Faster R-CNN on our custom dataset. However, Faster R-CNN struggled with small and medium-sized objects even after fine-tuning.

For saliency extraction, we explored the UniSal and SpectralResidual methods, generating saliency maps for each image using ground truth fixation maps from an eyetracker. Both methods yielded satisfactory results, with UniSal showing potential for improvement through training on eyetracker-acquired data in diverse traffic scenarios.

Our proposed method for billboard significance classification was evaluated, revealing better performance on saliency maps generated using the UniSal method. 
The method has delivered promising and satisfactory results in the classification of billboard significance. The proposed method achieved an accuracy of 82.6\% and a sensitivity of 74.0\%. This affirms the effectiveness of our approach in addressing the task at hand, marking a significant step forward in the field of billboard detection and saliency analysis. These findings contribute valuable insights to the ongoing development of methods for accurate and efficient billboard classification in various real-world scenarios. 

 \begin{acks} 
 This publication was supported by projects: (ACCORD, ITMS2014+: 313021X329), co-funded by the European Regional Development Fund, the TERAIS project, Horizon-Wider-2021 programme of the European Union under grant agreement number 101079338, and project APVV-23-0250.
\end{acks}

\bibliographystyle{ACM-Reference-Format}
\bibliography{sample-base}


\begin{thebibliography}{35}


\ifx \showCODEN    \undefined \def \showCODEN     #1{\unskip}     \fi
\ifx \showDOI      \undefined \def \showDOI       #1{#1}\fi
\ifx \showISBNx    \undefined \def \showISBNx     #1{\unskip}     \fi
\ifx \showISBNxiii \undefined \def \showISBNxiii  #1{\unskip}     \fi
\ifx \showISSN     \undefined \def \showISSN      #1{\unskip}     \fi
\ifx \showLCCN     \undefined \def \showLCCN      #1{\unskip}     \fi
\ifx \shownote     \undefined \def \shownote      #1{#1}          \fi
\ifx \showarticletitle \undefined \def \showarticletitle #1{#1}   \fi
\ifx \showURL      \undefined \def \showURL       {\relax}        \fi
\providecommand\bibfield[2]{#2}
\providecommand\bibinfo[2]{#2}
\providecommand\natexlab[1]{#1}
\providecommand\showeprint[2][]{arXiv:#2}

\bibitem[Abraham and Kovoor(2023)]%
        {abraham2023visual}
\bibfield{author}{\bibinfo{person}{Shilpa~Elsa Abraham} {and} \bibinfo{person}{Binsu~C Kovoor}.} \bibinfo{year}{2023}\natexlab{}.
\newblock \showarticletitle{Visual Saliency Modeling with Deep Learning: A Comprehensive Review}.
\newblock \bibinfo{journal}{\emph{Journal of Information \& Knowledge Management}} \bibinfo{volume}{22}, \bibinfo{number}{02} (\bibinfo{year}{2023}), \bibinfo{pages}{2250066}.
\newblock


\bibitem[Chan and Singhal(2013)]%
        {Chan13}
\bibfield{author}{\bibinfo{person}{Michelle Chan} {and} \bibinfo{person}{Anthony Singhal}.} \bibinfo{year}{2013}\natexlab{}.
\newblock \showarticletitle{The emotional side of cognitive distraction: Implications for road safety}.
\newblock \bibinfo{journal}{\emph{Accident Analysis {\&} Prevention}}  \bibinfo{volume}{50} (\bibinfo{year}{2013}), \bibinfo{pages}{147--154}.
\newblock


\bibitem[Chavan et~al\mbox{.}(2021)]%
        {chavanbillboard}
\bibfield{author}{\bibinfo{person}{Miss Sayali~Avinash Chavan}, \bibinfo{person}{Dermot Kerr}, \bibinfo{person}{Sonya Coleman}, {and} \bibinfo{person}{Mr~Hussein Khader}.} \bibinfo{year}{2021}\natexlab{}.
\newblock \showarticletitle{Billboard Detection in the Wild}. In \bibinfo{booktitle}{\emph{Irish Machine Vision and Image Conference Proceedings 2021}}. Irish Pattern Recognition and Classification Society., \bibinfo{pages}{57--64}.
\newblock


\bibitem[Costa et~al\mbox{.}(2019)]%
        {Costa19}
\bibfield{author}{\bibinfo{person}{Marco Costa}, \bibinfo{person}{Leonardo Bonetti}, \bibinfo{person}{Valeria Vignali}, \bibinfo{person}{Arianna Bichicchi}, \bibinfo{person}{Claudio Lantieri}, {and} \bibinfo{person}{Andrea Simone}.} \bibinfo{year}{2019}\natexlab{}.
\newblock \showarticletitle{Driver's visual attention to different categories of roadside advertising signs}.
\newblock \bibinfo{journal}{\emph{Applied ergonomics}}  \bibinfo{volume}{78} (\bibinfo{year}{2019}), \bibinfo{pages}{127--136}.
\newblock


\bibitem[Crundall et~al\mbox{.}(2006)]%
        {Crundall06}
\bibfield{author}{\bibinfo{person}{David Crundall}, \bibinfo{person}{Editha Van~Loon}, {and} \bibinfo{person}{Geoffrey Underwood}.} \bibinfo{year}{2006}\natexlab{}.
\newblock \showarticletitle{Attraction and distraction of attention with roadside advertisements}.
\newblock \bibinfo{journal}{\emph{Accident Analysis {\&} Prevention}} \bibinfo{volume}{38}, \bibinfo{number}{4} (\bibinfo{year}{2006}), \bibinfo{pages}{671--677}.
\newblock


\bibitem[Droste et~al\mbox{.}(2020)]%
        {droste2020unified}
\bibfield{author}{\bibinfo{person}{Richard Droste}, \bibinfo{person}{Jianbo Jiao}, {and} \bibinfo{person}{J~Alison Noble}.} \bibinfo{year}{2020}\natexlab{}.
\newblock \showarticletitle{Unified image and video saliency modeling}. In \bibinfo{booktitle}{\emph{Computer Vision--ECCV 2020: 16th European Conference, Glasgow, UK, August 23--28, 2020, Proceedings, Part V 16}}. Springer, \bibinfo{pages}{419--435}.
\newblock


\bibitem[Harasimczuk et~al\mbox{.}(2021)]%
        {Harasimczuk21}
\bibfield{author}{\bibinfo{person}{Justyna Harasimczuk}, \bibinfo{person}{Norbert~E Maliszewski}, \bibinfo{person}{Anna Olejniczak-Serowiec}, {and} \bibinfo{person}{Adam Tarnowski}.} \bibinfo{year}{2021}\natexlab{}.
\newblock \showarticletitle{Are longer advertising slogans more dangerous? The influence of the length of ad slogans on drivers’ attention and motor behavior}.
\newblock \bibinfo{journal}{\emph{Current Psychology}}  \bibinfo{volume}{40} (\bibinfo{year}{2021}), \bibinfo{pages}{429--441}.
\newblock


\bibitem[He et~al\mbox{.}(2017)]%
        {MaskRCNN}
\bibfield{author}{\bibinfo{person}{Kaiming He}, \bibinfo{person}{Georgia Gkioxari}, \bibinfo{person}{Piotr Doll{\'a}r}, {and} \bibinfo{person}{Ross Girshick}.} \bibinfo{year}{2017}\natexlab{}.
\newblock \showarticletitle{Mask r-cnn}. In \bibinfo{booktitle}{\emph{Proceedings of the IEEE International Conference on Computer Vision}}. IEEE, \bibinfo{pages}{2961--2969}.
\newblock


\bibitem[Hossari et~al\mbox{.}(2018)]%
        {hossari2018adnet}
\bibfield{author}{\bibinfo{person}{Murhaf Hossari}, \bibinfo{person}{Soumyabrata Dev}, \bibinfo{person}{Matthew Nicholson}, \bibinfo{person}{Killian McCabe}, \bibinfo{person}{Atul Nautiyal}, \bibinfo{person}{Clare Conran}, \bibinfo{person}{Jian Tang}, \bibinfo{person}{Wei Xu}, {and} \bibinfo{person}{Fran{\c{c}}ois Piti{\'e}}.} \bibinfo{year}{2018}\natexlab{}.
\newblock \showarticletitle{ADNet: A deep network for detecting adverts}.
\newblock \bibinfo{journal}{\emph{arXiv preprint arXiv:1811.04115}} (\bibinfo{year}{2018}).
\newblock


\bibitem[Hou and Zhang(2007)]%
        {hou2007saliency}
\bibfield{author}{\bibinfo{person}{Xiaodi Hou} {and} \bibinfo{person}{Liqing Zhang}.} \bibinfo{year}{2007}\natexlab{}.
\newblock \showarticletitle{Saliency detection: A spectral residual approach}. In \bibinfo{booktitle}{\emph{2007 IEEE Conference on computer vision and pattern recognition}}. Ieee, \bibinfo{pages}{1--8}.
\newblock


\bibitem[Itti et~al\mbox{.}(1998)]%
        {itti1998model}
\bibfield{author}{\bibinfo{person}{Laurent Itti}, \bibinfo{person}{Christof Koch}, {and} \bibinfo{person}{Ernst Niebur}.} \bibinfo{year}{1998}\natexlab{}.
\newblock \showarticletitle{A model of saliency-based visual attention for rapid scene analysis}.
\newblock \bibinfo{journal}{\emph{IEEE Transactions on pattern analysis and machine intelligence}} \bibinfo{volume}{20}, \bibinfo{number}{11} (\bibinfo{year}{1998}), \bibinfo{pages}{1254--1259}.
\newblock


\bibitem[Jiang et~al\mbox{.}(2015)]%
        {jiang2015salicon}
\bibfield{author}{\bibinfo{person}{Ming Jiang}, \bibinfo{person}{Shengsheng Huang}, \bibinfo{person}{Juanyong Duan}, {and} \bibinfo{person}{Qi Zhao}.} \bibinfo{year}{2015}\natexlab{}.
\newblock \showarticletitle{Salicon: Saliency in context}. In \bibinfo{booktitle}{\emph{Proceedings of the IEEE conference on computer vision and pattern recognition}}. \bibinfo{pages}{1072--1080}.
\newblock


\bibitem[Judd et~al\mbox{.}(2009)]%
        {judd2009learning}
\bibfield{author}{\bibinfo{person}{Tilke Judd}, \bibinfo{person}{Krista Ehinger}, \bibinfo{person}{Fr{\'e}do Durand}, {and} \bibinfo{person}{Antonio Torralba}.} \bibinfo{year}{2009}\natexlab{}.
\newblock \showarticletitle{Learning to predict where humans look}. In \bibinfo{booktitle}{\emph{2009 IEEE 12th international conference on computer vision}}. IEEE, \bibinfo{pages}{2106--2113}.
\newblock


\bibitem[Law and Deng(2018)]%
        {CornerNet}
\bibfield{author}{\bibinfo{person}{Hei Law} {and} \bibinfo{person}{Jia Deng}.} \bibinfo{year}{2018}\natexlab{}.
\newblock \showarticletitle{Cornernet: Detecting objects as paired keypoints}. In \bibinfo{booktitle}{\emph{Proceedings of the European Conference on Computer Vision}}. \bibinfo{pages}{734--750}.
\newblock


\bibitem[Lin et~al\mbox{.}(2017)]%
        {RetinaNet}
\bibfield{author}{\bibinfo{person}{Tsung-Yi Lin}, \bibinfo{person}{Priya Goyal}, \bibinfo{person}{Ross Girshick}, \bibinfo{person}{Kaiming He}, {and} \bibinfo{person}{Piotr Doll{\'a}r}.} \bibinfo{year}{2017}\natexlab{}.
\newblock \showarticletitle{Focal loss for dense object detection}. In \bibinfo{booktitle}{\emph{Proceedings of the IEEE international conference on computer vision}}. IEEE, \bibinfo{pages}{2980--2988}.
\newblock


\bibitem[Lin et~al\mbox{.}(2014)]%
        {MSCOCO}
\bibfield{author}{\bibinfo{person}{Tsung-Yi Lin}, \bibinfo{person}{Michael Maire}, \bibinfo{person}{Serge Belongie}, \bibinfo{person}{James Hays}, \bibinfo{person}{Pietro Perona}, \bibinfo{person}{Deva Ramanan}, \bibinfo{person}{Piotr Doll{\'a}r}, {and} \bibinfo{person}{C~Lawrence Zitnick}.} \bibinfo{year}{2014}\natexlab{}.
\newblock \showarticletitle{Microsoft coco: Common objects in context}. In \bibinfo{booktitle}{\emph{European conference on computer vision}}. Springer, \bibinfo{pages}{740--755}.
\newblock


\bibitem[Liu et~al\mbox{.}(2018)]%
        {Liu2018}
\bibfield{author}{\bibinfo{person}{Gang Liu}, \bibinfo{person}{Chuyi Wang}, {and} \bibinfo{person}{Yanzhong Hu}.} \bibinfo{year}{2018}\natexlab{}.
\newblock \showarticletitle{RPN with the Attention-based Multi-Scale Method and the Adaptive Non-Maximum Suppression for Billboard Detection}. In \bibinfo{booktitle}{\emph{2018 IEEE 4th International Conference on Computer and Communications (ICCC)}}. \bibinfo{pages}{1541--1545}.
\newblock
\urldef\tempurl%
\url{https://doi.org/10.1109/CompComm.2018.8780907}
\showDOI{\tempurl}


\bibitem[Maliszewski et~al\mbox{.}(2019)]%
        {Maliszewski19}
\bibfield{author}{\bibinfo{person}{Norbert Maliszewski}, \bibinfo{person}{Anna Olejniczak-Serowiec}, {and} \bibinfo{person}{Justyna Harasimczuk}.} \bibinfo{year}{2019}\natexlab{}.
\newblock \showarticletitle{Influence of sexual appeal in roadside advertising on drivers' attention and driving behavior}.
\newblock \bibinfo{journal}{\emph{PloS one}} \bibinfo{volume}{14}, \bibinfo{number}{5} (\bibinfo{year}{2019}), \bibinfo{pages}{e0216919}.
\newblock
\showISSN{1932-6203}


\bibitem[Marciano et~al\mbox{.}(2017)]%
        {Marciano17}
\bibfield{author}{\bibinfo{person}{Hadas Marciano} {et~al\mbox{.}}} \bibinfo{year}{2017}\natexlab{}.
\newblock \showarticletitle{The effect of billboard design specifications on driving: a pilot study}.
\newblock \bibinfo{journal}{\emph{Accident Analysis {\&} Prevention}}  \bibinfo{volume}{104} (\bibinfo{year}{2017}), \bibinfo{pages}{174--184}.
\newblock


\bibitem[Meuleners et~al\mbox{.}(2020)]%
        {Meuleners20}
\bibfield{author}{\bibinfo{person}{Lynn Meuleners}, \bibinfo{person}{Paul Roberts}, {and} \bibinfo{person}{Michelle Fraser}.} \bibinfo{year}{2020}\natexlab{}.
\newblock \showarticletitle{Identifying the distracting aspects of electronic advertising billboards: A driving simulation study}.
\newblock \bibinfo{journal}{\emph{Accident Analysis {\&} Prevention}}  \bibinfo{volume}{145} (\bibinfo{year}{2020}), \bibinfo{pages}{105710}.
\newblock


\bibitem[Mollu et~al\mbox{.}(2018)]%
        {Mollu18}
\bibfield{author}{\bibinfo{person}{Kristof Mollu}, \bibinfo{person}{Joris Cornu}, \bibinfo{person}{Kris Brijs}, \bibinfo{person}{Ali Pirdavani}, {and} \bibinfo{person}{Tom Brijs}.} \bibinfo{year}{2018}\natexlab{}.
\newblock \showarticletitle{Driving simulator study on the influence of digital illuminated billboards near pedestrian crossings}.
\newblock \bibinfo{journal}{\emph{Transportation research part F: traffic psychology and behaviour}}  \bibinfo{volume}{59} (\bibinfo{year}{2018}), \bibinfo{pages}{45--56}.
\newblock


\bibitem[Neuhold et~al\mbox{.}(2017)]%
        {neuhold2017mapillary}
\bibfield{author}{\bibinfo{person}{Gerhard Neuhold}, \bibinfo{person}{Tobias Ollmann}, \bibinfo{person}{Samuel Rota~Bulo}, {and} \bibinfo{person}{Peter Kontschieder}.} \bibinfo{year}{2017}\natexlab{}.
\newblock \showarticletitle{The mapillary vistas dataset for semantic understanding of street scenes}. In \bibinfo{booktitle}{\emph{Proceedings of the IEEE international conference on computer vision}}. \bibinfo{pages}{4990--4999}.
\newblock


\bibitem[Rahmat et~al\mbox{.}(2019)]%
        {rahmat2019advertisement}
\bibfield{author}{\bibinfo{person}{Romi~Fadillah Rahmat}, \bibinfo{person}{Dennis Dennis}, \bibinfo{person}{Opim~Salim Sitompul}, \bibinfo{person}{Sarah Purnamawati}, {and} \bibinfo{person}{Rahmat Budiarto}.} \bibinfo{year}{2019}\natexlab{}.
\newblock \showarticletitle{Advertisement billboard detection and geotagging system with inductive transfer learning in deep convolutional neural network}.
\newblock \bibinfo{journal}{\emph{TELKOMNIKA (Telecommunication Computing Electronics and Control)}} \bibinfo{volume}{17}, \bibinfo{number}{5} (\bibinfo{year}{2019}), \bibinfo{pages}{2659--2666}.
\newblock


\bibitem[Redmon et~al\mbox{.}(2016)]%
        {Yolo}
\bibfield{author}{\bibinfo{person}{Joseph Redmon}, \bibinfo{person}{Santosh Divvala}, \bibinfo{person}{Ross Girshick}, {and} \bibinfo{person}{Ali Farhadi}.} \bibinfo{year}{2016}\natexlab{}.
\newblock \showarticletitle{You only look once: Unified, real-time object detection}. In \bibinfo{booktitle}{\emph{Proceedings of the IEEE conference on computer vision and pattern recognition}}. IEEE, \bibinfo{pages}{779--788}.
\newblock


\bibitem[Ren et~al\mbox{.}(2015)]%
        {FasterRCNN}
\bibfield{author}{\bibinfo{person}{Shaoqing Ren}, \bibinfo{person}{Kaiming He}, \bibinfo{person}{Ross Girshick}, {and} \bibinfo{person}{Jian Sun}.} \bibinfo{year}{2015}\natexlab{}.
\newblock \showarticletitle{Faster r-cnn: Towards real-time object detection with region proposal networks}. In \bibinfo{booktitle}{\emph{Advances in neural information processing systems}}. \bibinfo{pages}{91--99}.
\newblock


\bibitem[Riche et~al\mbox{.}(2013)]%
        {Riche-13}
\bibfield{author}{\bibinfo{person}{N. Riche}, \bibinfo{person}{M. Duvinage}, \bibinfo{person}{M. Mancas}, \bibinfo{person}{B. Gosselin}, {and} \bibinfo{person}{T. Dutoit}.} \bibinfo{year}{2013}\natexlab{}.
\newblock \showarticletitle{Saliency and human fixations: State-of-the-art and study of comparison metrics}. In \bibinfo{booktitle}{\emph{International Conference on Computer Vision}}. \bibinfo{publisher}{IEEE}, \bibinfo{pages}{1153--1160}.
\newblock


\bibitem[Sandler et~al\mbox{.}(2018)]%
        {sandler2018mobilenetv2}
\bibfield{author}{\bibinfo{person}{Mark Sandler}, \bibinfo{person}{Andrew Howard}, \bibinfo{person}{Menglong Zhu}, \bibinfo{person}{Andrey Zhmoginov}, {and} \bibinfo{person}{Liang-Chieh Chen}.} \bibinfo{year}{2018}\natexlab{}.
\newblock \showarticletitle{Mobilenetv2: Inverted residuals and linear bottlenecks}. In \bibinfo{booktitle}{\emph{Proceedings of the IEEE conference on computer vision and pattern recognition}}. \bibinfo{pages}{4510--4520}.
\newblock


\bibitem[Sun and Fisher(2003)]%
        {sun2003object}
\bibfield{author}{\bibinfo{person}{Yaoru Sun} {and} \bibinfo{person}{Robert Fisher}.} \bibinfo{year}{2003}\natexlab{}.
\newblock \showarticletitle{Object-based visual attention for computer vision}.
\newblock \bibinfo{journal}{\emph{Artificial intelligence}} \bibinfo{volume}{146}, \bibinfo{number}{1} (\bibinfo{year}{2003}), \bibinfo{pages}{77--123}.
\newblock


\bibitem[Tarnowski et~al\mbox{.}(2017)]%
        {Tarnowski17}
\bibfield{author}{\bibinfo{person}{Adam Tarnowski}, \bibinfo{person}{Anna Olejniczak-Serowiec}, {and} \bibinfo{person}{Agnieszka Marszalec}.} \bibinfo{year}{2017}\natexlab{}.
\newblock \showarticletitle{Roadside advertising and the distraction of driver’s attention}. In \bibinfo{booktitle}{\emph{MATEC Web of Conferences}}, Vol.~\bibinfo{volume}{122}. EDP Sciences, \bibinfo{pages}{03010}.
\newblock


\bibitem[Tian et~al\mbox{.}(2019)]%
        {tian2019fcos}
\bibfield{author}{\bibinfo{person}{Zhi Tian}, \bibinfo{person}{Chunhua Shen}, \bibinfo{person}{Hao Chen}, {and} \bibinfo{person}{Tong He}.} \bibinfo{year}{2019}\natexlab{}.
\newblock \showarticletitle{Fcos: Fully convolutional one-stage object detection}. In \bibinfo{booktitle}{\emph{Proceedings of the IEEE/CVF international conference on computer vision}}. \bibinfo{pages}{9627--9636}.
\newblock


\bibitem[Ullah et~al\mbox{.}(2020)]%
        {ullah2020brief}
\bibfield{author}{\bibinfo{person}{Inam Ullah}, \bibinfo{person}{Muwei Jian}, \bibinfo{person}{Sumaira Hussain}, \bibinfo{person}{Jie Guo}, \bibinfo{person}{Hui Yu}, \bibinfo{person}{Xing Wang}, {and} \bibinfo{person}{Yilong Yin}.} \bibinfo{year}{2020}\natexlab{}.
\newblock \showarticletitle{A brief survey of visual saliency detection}.
\newblock \bibinfo{journal}{\emph{Multimedia Tools and Applications}}  \bibinfo{volume}{79} (\bibinfo{year}{2020}), \bibinfo{pages}{34605--34645}.
\newblock


\bibitem[Vig et~al\mbox{.}(2014)]%
        {vig2014large}
\bibfield{author}{\bibinfo{person}{Eleonora Vig}, \bibinfo{person}{Michael Dorr}, {and} \bibinfo{person}{David Cox}.} \bibinfo{year}{2014}\natexlab{}.
\newblock \showarticletitle{Large-scale optimization of hierarchical features for saliency prediction in natural images}. In \bibinfo{booktitle}{\emph{Proceedings of the IEEE conference on computer vision and pattern recognition}}. \bibinfo{pages}{2798--2805}.
\newblock


\bibitem[Zalesinska(2018)]%
        {Zalesinska18}
\bibfield{author}{\bibinfo{person}{Malgorzata Zalesinska}.} \bibinfo{year}{2018}\natexlab{}.
\newblock \showarticletitle{The impact of the luminance, size and location of LED billboards on drivers’ visual performance—laboratory tests}.
\newblock \bibinfo{journal}{\emph{Accident Analysis {\&} Prevention}}  \bibinfo{volume}{117} (\bibinfo{year}{2018}), \bibinfo{pages}{439--448}.
\newblock


\bibitem[Zhang et~al\mbox{.}(2020)]%
        {zhang2020bridging}
\bibfield{author}{\bibinfo{person}{Shifeng Zhang}, \bibinfo{person}{Cheng Chi}, \bibinfo{person}{Yongqiang Yao}, \bibinfo{person}{Zhen Lei}, {and} \bibinfo{person}{Stan~Z Li}.} \bibinfo{year}{2020}\natexlab{}.
\newblock \showarticletitle{Bridging the gap between anchor-based and anchor-free detection via adaptive training sample selection}. In \bibinfo{booktitle}{\emph{Proceedings of the IEEE/CVF conference on computer vision and pattern recognition}}. \bibinfo{pages}{9759--9768}.
\newblock


\bibitem[Zhou et~al\mbox{.}(2019)]%
        {CenterNet}
\bibfield{author}{\bibinfo{person}{Xingyi Zhou}, \bibinfo{person}{Dequan Wang}, {and} \bibinfo{person}{Philipp Kr{\"a}henb{\"u}hl}.} \bibinfo{year}{2019}\natexlab{}.
\newblock \showarticletitle{Objects as points}.
\newblock \bibinfo{journal}{\emph{arXiv preprint arXiv:1904.07850}} (\bibinfo{year}{2019}).
\newblock


\end{thebibliography}


\end{document}